\documentclass[a4paper,10pt]{article}
\pdfoutput=1

\usepackage[T1]{fontenc}
\usepackage[utf8]{inputenc}
\usepackage{geometry}
\usepackage[english]{babel}
\usepackage{moreverb}
\usepackage{latexsym}
\usepackage{todonotes}
\usepackage{bbm}
\usepackage{graphicx}
\usepackage{tikz}
\usepackage{pgfplots}
\usepackage{epsfig}
\usepackage{amsmath}
\usepackage{authblk}
\usepackage{amsfonts}
\usepackage{amssymb}
\usepackage{pifont}
\usepackage{calc}
\usepackage{time}
\usepackage{animate}
\usepackage{subfigure}
\usepackage{psfrag}
\usepackage{dsfont}
\usepackage{array}
\usepackage{cases}
\usepackage{color}
\usepackage{xparse}
\usepackage{breqn}
\usepackage{url}
\usepackage{hyperref}
\usepackage{listings}
\usepackage{float}
\usepackage{cancel}

\newcommand{\R}{\mathbb R}

\newtheorem{theorem}{Theorem}[section]
\newtheorem{proposition}{Proposition}[section]

\newcommand{\tbf}[1]{\textbf{#1}}
\newcommand{\diff}[2]{\frac{\partial #1}{\partial #2}}

\title{A massively parallel multi-level approach to a domain decomposition
  method for the optical flow estimation with varying illumination}
\date{}
\author[1]{Diane Gilliocq-Hirtz\footnote{LMIA, 6 rue des Frères Lumière,
    68093 Mulhouse, France.\\
    Email : diane.gilliocq-hirtz@uha.fr}
\and Zakaria Belhachmi\footnote{LMIA, 6 rue des Frères Lumière,
  68093 Mulhouse, France\\
  Email : zakaria.belhachmi@uha.fr}}

\usetikzlibrary{patterns}
\begin{document}
\maketitle


\begin{abstract}
We consider a variational method to solve the optical flow problem with varying
illumination. We apply an adaptive control of the regularization parameter which
allows us to preserve the edges and fine features of the computed flow. To reduce
the complexity of the estimation for high resolution images and the time of computations, 
we implement a multi-level parallel approach based on the domain decomposition with the 
Schwarz overlapping method. The second level of
parallelism uses the massively parallel solver \emph{MUMPS}. We perform some
numerical simulations to show the efficiency of our approach and to validate it on
classical and real-world image sequences.
\end{abstract}


\section*{Introduction}

In the last decades, the estimation of the optical flow has become a
popular and central problem in computer vision (\cite{Weickertvariational1}, \cite{bruhn-clg-2005},
\cite{schnorr1}, \cite{deriche}). It is involved, for example, in almost all the 
movies and pictures compression processes, or in the obstacle detection in the new
smart cars and in robotics. The modelling and the detection of 
the motion in a scene involve numerous difficulties (e.g. the aperture problem, 
occlusions, \dots \cite{Weickertvariational1}).
Among such difficulties, for the optical flow 
estimation, one with growing importance is the cost of the method in term of 
computation time (and the storage), which rises with the increasing resolution of the images due 
the technological devices progress.
Up to now, there exist numerous methods for the optical flow estimation among which the Partial 
Differential Equations and particularly, the variational methods turn to be very efficient. 
They offer a complete framework which consists of mathematically founded continuous models, 
and a large number of numerical methods (\cite{Weickertvariational1}, \cite{bruhn-clg-2005}, 
\cite{schnorr1}, \cite{deriche}). 
They allow to cope with the ill-posedness of the optical flow
problem, due to the aperture problem \cite{bruhn-clg-2005}, by including a large range of 
regularization procedures.
In this article, we consider a variational method based on the linear Horn and Schunck approach 
\cite{horn-global-1981} but with a variable regularization parameter. This method introduced in 
\cite{belhachmi-regu-2010} is proved to be an efficient approach to solve the optical flow problem in the 
sense that it is edge preserving and low cost (in term of degrees of freedom) \cite{belhachmi-segmentation-2014}. 

A large part of the research works in the literature deals with the determination of optical flow 
under an assumption of constant illumination. 
This assumption is not satisfied in general and becomes a
source of inaccurate estimation and serious limitations in many applications. Besides, modelling a 
varying illumination is quite complex and increases the ill-posed character of the problem. 
In this article, following Gennert and Negahdaripour \cite{gennert-illu}, we assume given an a priori law 
for the illumination variations and we 
introduce a supplementary variable, which models the illumination, in the system of equations. 
We show that the obtained partial differential equations system solved in the framework of the adaptive variational 
approach \cite{belhachmi-regu-2010} is an efficient and innovative method for the optical flow estimation with 
varying illumination. 

A classical criticism against the variational methods is their "complexity" and
 their potential time consuming character, particularly when using 
unstructured meshes and finite element method discretization. 
The main contribution of this article is to propose and to validate an efficient 
massively parallel multi-level solver using domain decomposition method to solve 
the optical flow problem with varying illumination, following the variational adaptive approach 
proposed in \cite{belhachmi-regu-2010}. 
We obtain the optical flow with an accurate estimation and a significant reduction of the time of computations. 
We validate our method on a classical benchmark and two real-world image sequences. 

In Section 1, we recall the Horn and Schunck model for the optical flow estimation and we give the system of 
equations to solve in the case where we allow illumination variations. 
In Section 2, we
briefly recall the finite element method and we will rewrite the
discrete optical flow system. We also define the adaptive strategy and give the resulting algorithm.
In Section 3, we will consider the domain decomposition method with overlapping and the additive
Schwarz method used to solve the algebraic problem. The Section 4 concerns the numerical simulations. 
We present some
results obtained with our method, in particular, we give in detail the optical flow estimation for the 
so-called RubberWhale sequence to show the performances of the approach. We also perform the analysis of
the computation time of the massively parallel algorithm. Finally, we will present two examples of the 
computation of the optical flow for real-world sequences.

\section{Optical flow problem}

\begin{figure}[H]
  \begin{center}
    \begin{tikzpicture}{scale=1}
      \draw (0,1) -- (2,1);
      \draw (1,1) circle (20pt);
      \draw[red,->] (1,1) -- (1,1.5);
      \draw (1,-0.5) node {$1$};

      \draw (4,2) -- (4,0);
      \draw (4,1) circle (20pt);
      \draw[red,->] (4,1) -- (4.5,1);
      \draw (4,-0.5) node {$2$};

      \draw (6,2) -- (8,2);
      \draw (8,2) -- (8,0);
      \draw (8,2) circle (20pt);
      \draw[red,->] (8,2) -- (8.3,2.3);
      \draw (7,-0.5) node {$3$};
    \end{tikzpicture}
  \end{center}
  \caption{Representation of the aperture problem.}
  \label{fig:aperture}
\end{figure}
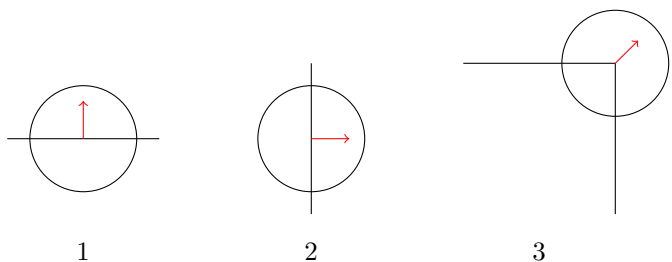

We consider a sequence of two successive frames where $\Omega\subset\R^2$ is the image domain. 
The intensity of a pixel $(x,y)$ at an instant $t$ is
defined by the function 
\begin{equation*}
  \begin{tabular}{llll}
    I: & $\Omega\times[0,T]$ & $\rightarrow$ & $\R$\\
       & $(x,y),t$ & $\mapsto$ & $I(x,y,t).$
  \end{tabular}
\end{equation*}
As it is usually done \cite{barron-smooth-1994}, we use a convolution with a
Gaussian kernel $K_\sigma$ of standard deviation $\sigma$ to work with smoothed
images. We define the smoothed image sequence by
\begin{equation*}
  f(x,y,t) = (K_\sigma\star I)(x,y,t).
\end{equation*}
The optical flow $\tbf u=(u_1,u_2)$ represents the vector field describing the motion of each
pixel in the sequence. 
Following Horn and Schunck \cite{horn-global-1981}, we first consider the
brightness constancy assumption. It means that the brightness intensity stays the same
between two successive frames,
\begin{equation}
  \label{eq:brightness}
  f(x,y,t)=f(x+u_1,y+u_2,t+1)
\end{equation}
which is equivalent to say that the variation in time is null
\begin{equation*}
  \frac{\partial f}{\partial t}(x,y,t)=0.
\end{equation*}
Since the displacements are supposed to be small, we can assume that $f$ is $\mathcal C^1( \left[ 0,\infty
\right[ ;\R)$. So by using a first order Taylor expansion, we obtain the
fundamental constraint of the optical flow
\begin{equation}
  \label{eq:dfdt}
  \frac{\partial f}{\partial t}+\frac{\partial f}{\partial x}\frac{\partial
    x}{\partial t}+\frac{\partial f}{\partial y}\frac{\partial y}{\partial
    t}=0.
\end{equation}
The time derivatives of $x$ and $y$ represent the component $u_1$ and $u_2$ of
the optical flow. Hence, with the notation $f_*=\diff{f}{*}$ the equation
(\ref{eq:dfdt}) becomes

\begin{equation}
  \label{eq:ofeq}
  f_xu_1+f_yu_2+f_t=0.
\end{equation}
In this way, we have to determine two unknowns $u_1$ and $u_2$ with only one
equation. This is the so-called \emph{aperture problem} illustated in the figure
\ref{fig:aperture}. In the example $1$, in a
local neighborhood we can only detect the vertical motion. In the example $2$,
it is only the horizontal motion that we can find. The example $3$ is the only one
where we can detect a diagonal motion.

To go through this ill-posedness, Lucas and Kanade \cite{lucas-local-1981}
assume that the motion is constant in a neighborhood of size $\rho$.
Contrary to this local assumption, Horn and Schunck propose
\cite{horn-global-1981} a global approach to overcome the aperture
problem. They introduce a regularization parameter $\alpha$
which acts as a penalizer and leads to a smoother flow field (the bigger $\alpha$
is, the smoother the flow is). The idea of Bruhn and Weickert \cite{bruhn-clg-2005}
is, to combine both methods and minimize the functional

\begin{equation}
  \label{eq:BW}
  \int_{\Omega} K_{\rho}\star (f_xu_1+f_yu_2+f_t)^2+
  \alpha (|\nabla u_1|^2+|\nabla u_2|^2)dxdy
\end{equation}
where $K_{\rho}$ is a Gaussian deviation of parameter $\rho$ and $\alpha$ is a
constant regularization parameter.\\

On real-world images, the assumption of the brightness constancy is no longer
verified. Occlusions, shadows or glints don't meet this constraint. Hence, the estimation
of the previous model is not accurate so we need to consider another
assumption. Different approaches were proposed to model illumination variations,
such as the assmption of the constancy of the gradient amplitude \cite{bruhn-phd-2006} or, in the
case of color images, we can cite the work with variables less sensitive to such illumination
changes \cite{weickert-color-2007}. In this article, following Gennert and
Negahdaripour \cite{gennert-illu}, we consider a varying illumination obeying an
affine transformation. This assumption, even covering a wide range of
applications, might appear as a strong one. Moreover, it introduces a
supplementary unknown, enforcing the ill-posedness of the optical flow
estimation. To balance such potential shortcomings, we couple this modelling
with an adaptive control of the regularization of the parameter associated to the new
unknown. In this article, we restrict ourselves to smooth variations, keeping
the parameter large enough. The assumption allowing a linear motion of the
brightness intensity between the two images becomes

\begin{equation*}
  f(x+u_1,y+u_2,t+1)=M(x,y,t)f(x,y,t)+T(x,y,t)
\end{equation*}
where $M$ is the multiplier and $T$ is the translator. In our case, we suppose
that the translator is negligible. The smaller the displacement is, the closer to $1$
$M$ is. In this way, we can set

\begin{equation}
  \label{eq:deltam}
   M=1+\delta m
\end{equation}
where $\delta m$ tends to zero when the displacement is very
small. That gives the new equation of the optical flow 

\begin{equation}
  \label{eq:OF2}
  f_xu_1+f_yu_2+f_t-fm_t = 0
\end{equation}
where $m_t$ is the derivative of $M$. For more details, we refer the reader to
\cite{gennert-illu}. The optical flow problem allowing
varying illumination consists finally of minimizing the functional

\begin{equation}
  \label{eq:BW2}
  \int_{\Omega} K_{\rho}\star (f_xu_1+f_yu_2+f_t-m_tf)^2+
  \alpha (|\nabla u_1|^2+|\nabla u_2|^2) + \lambda |\nabla m_t|^2dxdy.
\end{equation}
According to Euler-Lagrange equations, we have the system

\begin{numcases}{}
  -\text{div}( \Lambda\nabla \tbf{U} ) + A_{\rho}\tbf{U} = \tbf{F} \text{ in }
  \Omega\label{eq:eqpro1illu}\\
  \diff{\tbf{U}}{n} = 0 \text{ on } \partial\Omega\nonumber
\end{numcases}
with

\begin{equation*}
  \Lambda = 
  \begin{bmatrix}
    \alpha\\
    \alpha\\
    \lambda
  \end{bmatrix}
  \text{, }
  \tbf{U} = 
  \begin{bmatrix}
    u_1\\
    u_2\\
    m_t
  \end{bmatrix}
  \text{, }
  A_{\rho} = 
  \begin{bmatrix}
    K_{\rho}\star (f_x)^2  & K_{\rho}\star (f_xf_y)  & -K_{\rho}\star (f_xf)\\
    K_{\rho}\star (f_yf_x) & K_{\rho}\star (f_y)^2   & -K_{\rho}\star (f_yf)\\
    -K_{\rho}\star (ff_x)  & -K_{\rho}\star (ff_y)   &  K_{\rho}\star (f)^2
  \end{bmatrix}
\end{equation*}
and

\begin{equation*}
  \tbf{F} = 
  \begin{bmatrix}
    -K_{\rho}\star (f_xf_t)\\
    -K_{\rho}\star (f_yf_t)\\
    K_{\rho}\star (ff_t)
  \end{bmatrix}.
\end{equation*}

It is well known that taking $\alpha$ and $\lambda$ constant, even well chosen,
leads to undesired oversmoothing and blurs the edges of an image. Thus,
following the idea of (\cite{belhachmi-regu-2010},
\cite{belhachmi-segmentation-2014}), we will consider the general setting where
$\alpha$ and $\lambda$ are discontinuous and piecewise constant functions in
order to prevent these smoothing effects.
Indeed, as proved in \cite{belhachmi-segmentation-2014}, choosing a small value
of $\alpha$ in regions where there are edges gives sharper edges, then an
improved restitution of the motion and its segmentation. We 
refer to \cite{belhachmi-segmentation-2014} for more details.
In particular, we assume given $\displaystyle\Omega = \mathop{\cup}_i\Omega_i$,
$\alpha = (\alpha_i)_{i\in I}$, $\lambda = (\lambda_i)_{i\in I}$, and
$\alpha_m$, the non-negative minimal value of $\alpha_\ell$. Then, we have the
following theorem.

\begin{theorem}
  The problem (\ref{eq:eqpro1illu}) has a unique solution in $H^1(\Omega)$.
\end{theorem}
The well-posedness is obtained from the Lax-Milgram lemma. 
By assuming that $\Omega$ is Lipschitz, we can work in the Hilbert space
$H^1(\Omega)$. We define the norm
\begin{equation*}
  \Vert \tbf U\Vert_{\rho,\Lambda}^2 = \Vert \Lambda^{\frac 1 2} \nabla \tbf U
  \Vert_{L^2(\Omega)}^2 + \Vert \tbf A_\rho^{\frac 1 2}\tbf U\Vert_{L^2(\Omega)}^2
\end{equation*}
where $\Vert \tbf A_\rho^{\frac 1 2}\tbf U\Vert_{L^2(\Omega)}^2 = (A_{\rho} \tbf
U, \tbf U)$.
Let denote $\beta_M = \max(\alpha_M,\lambda_M)$ and \\$\beta_m =
\min(\alpha_m,\lambda_m)$, a straightforward extension of
(\cite{belhachmi-segmentation-2014}, proposition 2.7) gives the following proposition.

\begin{proposition}
  Let $\tbf U$ be a solution of (\ref{eq:OF2}). 
  There exists $C>0$ independent of $\alpha$ such that for $\tbf U_{\Lambda}$, the
  solution of the optical flow problem where the regularization is a piecewise constant
  function, the following inequalities hold

  \begin{equation}
    \|\tbf U_{\Lambda}\|_{\rho, \Lambda} \leq C\|A_\rho^{\frac{1}{2}}\tbf
    U\|_{L^2(\Omega)}.
    \label{ineq1}
  \end{equation}
  and

  \begin{equation}
    \|\tbf U - \tbf U_{\Lambda}\|_{\rho, \Lambda} \leq
    C\left(\frac{\beta_M}{\beta_m}\right)^{\frac{1}{2}}\|\alpha^{\frac{1}{2}}
    \nabla\tbf U_\Lambda\|_{L^2(\Omega)}.
    \label{ineq2}
  \end{equation}
\end{proposition}
From now on, we denote $\tbf U$ in place of $\tbf U_\Lambda$ for brevity. 


\section{Finite elements implementation}

\subsection{Variational form}

In order to solve the system (\ref{eq:eqpro1illu}), we use the finite element
method. To do so, we need to write this equation under its variational
form. We multiply the first equation of (\ref{eq:eqpro1illu}) by a test
function $\boldsymbol\varphi$, we integrate over $\Omega$ and by using the Green formula on
the first integral, we have

\begin{equation}
\label{weakform}
  \int_\Omega \Lambda\nabla\tbf U\nabla\boldsymbol\varphi d\tbf x
   + \int_\Omega \tbf A_{\rho}\tbf{U}\boldsymbol\varphi d\tbf x =
  \int_\Omega \tbf F\boldsymbol\varphi d\tbf x, \qquad \forall
  \boldsymbol\varphi\in\mathcal (H^1(\Omega))^3. 
\end{equation}

\subsection{Discretization}

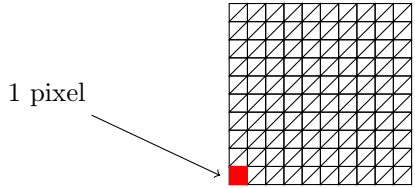
\begin{figure}[H]
  \centering
  \begin{tikzpicture}[scale=0.24]
    \foreach \i in {0,...,9}{
      \foreach \j in {0,...,9}{
        \draw (\i+0,\j+0) -- (\i+1,\j+0) -- (\i+1,\j+1) -- (\i+0,\j+0) --
        (\i+0,\j+1) -- (\i+1,\j+1);
      }
    }
    \fill[red] (0,0) rectangle (1,1);
    \draw (-10,5) node (A) {1 pixel};
    \draw[->] (A) -- (-0.5,0.5);
  \end{tikzpicture}
  \caption{Representation of the initial mesh $\mathcal T_h$. Each square
    represents a pixel of the image and consists of two triangles.}
  \label{fig:sch-mesh}
\end{figure}

We consider a regular triangular mesh $\mathcal T_h$ (figure \ref{fig:sch-mesh}). The estimation consists of
solving a linear problem on each cell of the mesh.
We define the space of approximations by

\begin{equation*}
  \mathcal V_h = \{\mathbf V_h\in C(\bar\Omega),\quad \mathbf V_h|_K\in P_1(K)^3\}
\end{equation*}
where $P_1(K)$ is the space of the linear functions on $K\in\mathcal T_h$. If we
set $A_{\rho,h}$ a finite element approximation of $A_\rho$, the discrete
problem of the optical flow states
\begin{equation}
  \label{eq:discret-form}
  \left\{
    \begin{tabular}{l}
      \text{Find $\mathbf{U}_{h}\in \mathcal V_h^3$, such that}\\
      \text{$\displaystyle\int_{\mathcal T_h}\Lambda\nabla\mathbf{U}_{h}\cdot\nabla 
      \mathbf{V}_{h}dxdy +
      \int_{\mathcal T_h}\mathbf{V}_{h}^T\tbf A_{\rho,h}\mathbf{U}_{h}dxdy=
      \int_{\mathcal T_h}\mathbf{F}_{h}\cdot\mathbf{V}_{h}dxdy,\quad\forall
      \mathbf{V}_{h}\in \mathcal V_h^3$.}
    \end{tabular}
  \right.
\end{equation}
We can show, by using the Lax-Milgram lemma that this weak formulation
admits a unique solution.


\subsection{Adaptive regularization}

In this part, we are interested in controlling the parameter $\alpha$. The
regularization parameter is now considered as a piecewise constant function. This local
choice of $\alpha$ is based on an a posteriori strategy analysis and was
proposed by Belhachmi and Hecht \cite{belhachmi-regu-2010}. 
The choice of the regularization is motivated by the fact that a small value is
usefull to correctly approximate the Neumann boundary conditions on the edges of
objects. However, it increases the maximum value of the optical flow. So, in order to
have a better estimation, we prefer a large regularization. The local choice of
$\alpha$ allows to decrease its value on regions where we need a small $\alpha$
and keep a large value in the rest of the image. In our case, we have the
additional unknown $m_t$ with its regularization parameter $\lambda$ and in a first
time, we keep this parameter constant.\\
Since we want to locally choose the regularization parameter, we will have a
large ratio $\frac{\alpha_M}{\alpha_m}$, so we will use the inequality
(\ref{ineq2}) in the error indicator. The control of the regularization is done
through an error indicator which is given for each element $K\in\mathcal T_h$ by

\begin{dmath}
  \eta_K = \Lambda_K^{-\frac{1}{2}}h_K\|\tbf{F}_h +
  \text{div}( \Lambda_K\nabla\tbf{U}_{\Lambda,h} ) +
  A_{\rho,h}\tbf{U}_{\Lambda,h}\|_{L^2(K)^2} +
  \frac{1}{2}\sum_{e\in\varepsilon_K} \Lambda_e^{-\frac{1}{2}}h_e^{\frac{1}{2}}
  \|[\Lambda\nabla\tbf{U}_{\Lambda,h}\cdot\tbf{n}_e]_e\|_{L^2(e)^2}\label{errind}
\end{dmath}
where $\varepsilon_K$ represents the set of all
edges $e$ of $K$. The diameter of $K$ is noted $h_K$ and the diameter of an
edge $e$ is
$h_e$. $\tbf{n}_e$ represents the normal vector from $e$, $\Lambda_e$ is the maximum
between the $\Lambda$ of the two neighbors of an edge, and
$[.]_e$ represents the jump over the edge $e$ which means the difference between
the outside and inside values.
The error indicator $\eta_K$ describes the finite element error and the model
error. On the potential set of discontinuities, this value is large because $\nabla U_{\Lambda,h}$
is large. In fact, when the brightness is abruptly changing in an area,
it means that we are close to an edge for the optical flow. So, to improve the
solution, we decrease $\alpha$ from the two first components of $\Lambda$ and the
third component $\lambda$ stays constant. The decreasing formula for $\alpha$
is given by

\begin{equation*}
  \alpha_K^{n+1} = \max\left( \frac{\alpha_K^n}{1 +
      \kappa\max\left(\frac{\eta_K}{\|\eta_K\|_\infty} - 
        \eta ,0\right)},\alpha_{th} \right)
\end{equation*}
where $\kappa$ is an arbitrary control parameter and $\alpha_{th}$ is a threshold. In this
way, if the relative error (measured with $\eta_K$) is greater than a given value $\eta$ we reduce in $K$, 
the value of $\alpha$. On the other hand, if it is less than $\eta$ the denominator is equal to
one and so $\alpha$ remains unchanged. This local and adaptive control of $\alpha$ is implemented with 
the algorithm 
\begin{enumerate}
    \item Compute of a first approximation $\tbf U_\alpha^0$ of the optical
        flow. This estimation is done on a cartesian grid $T_h^0$ where we
        have one cell per pixel. Define $i=0$.
    \item $i = i+1$. Build an adapted mesh $T_h^i$ with the metric error indicator.
    \item Update of $\alpha_i(x,y)$ on $T_h^i$.
    \item Go to step 2.
\end{enumerate}
The convergence of this algorithm when the mesh size goes to zero is proved in 
\cite{belhachmi-segmentation-2014}.


\section{Domain Decomposition}

\subsection{Image decomposition}

\begin{figure}[H]
  \centering
  \begin{tikzpicture}
    \node[anchor=south west,inner sep=0,opacity=0.65] at (0,0)
    {\includegraphics[width=5.2cm,height=4.8cm]{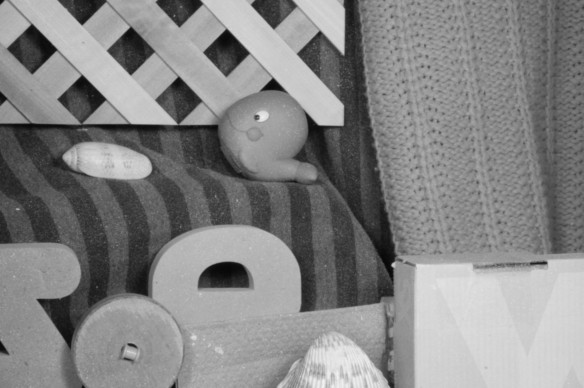}};
    \draw[red,very thick] (0,2.4)--(5.2,2.4);
    \draw[red,very thick] (2.6,0)--(2.6,4.8);
    \node[draw] at (-1,1.2) {CPU 0};
    \node[draw] at (6.2,1.2) {CPU 1};
    \node[draw] at (6.2,3.6) {CPU 3};
    \node[draw] at (-1,3.6) {CPU 2};
  \end{tikzpicture}
  \caption{Decomposition of an image for four CPU.}
  \label{fig:DDM1}
\end{figure}
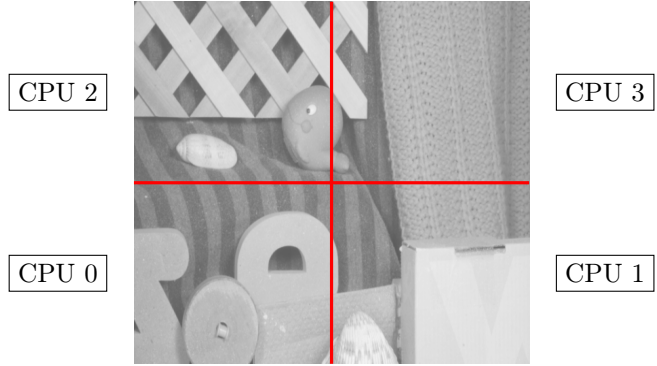

In the case where we want to estimate the optical flow between two
large images, we have implemented a domain decomposition. Indeed, as we have
seen on the figure \ref{fig:sch-mesh}, if we have large number of pixels on the
picture, we have a large number of triangles in the mesh. In the finite element
method, the resolution consists of solving a large system with a large number of
degrees of freedom. The aim is that every Central Processing Unit (CPU) computes the
optical flow of one part of the image (see figure \ref{fig:DDM1}). So, the number of
linear systems to solve is reduced by a factor equivalent to the number of CPU
used.
The domain decomposition method allows us to obtain performances which are
difficult to reach with classical variational methods and even worse with the
adaptive process to control the parameter $\alpha$.

\begin{figure}[H]
  \centering
  \begin{tikzpicture}
    \draw[very thin,gray] (0,0) grid[step=1./24] (2,2);
    \node at (1,-0.3) {ratio = 1.846};
    \draw[very thin,gray] (3,0) grid[step=1./24] (5,2);
    \node at (4,-0.3) {ratio = 3};
    \draw[very thin,gray] (6,0) grid[step=1./24] (8,2);
    \node at (7,-0.3) {ratio = 3.428};
    \draw[very thin,gray] (0,3) grid[step=1./24] (2,5);
    \node at (1,2.7) {ratio = 1.846};
    \draw[very thin,gray] (3,3) grid[step=1./24] (5,5);
    \node at (4,2.7) {ratio = 3};
    \draw[very thin,gray] (6,3) grid[step=1./24] (8,5);
    \node at (7,2.7) {ratio = 3.428};

    \draw[red,thick] (1./6,0)--(1./6,2);
    \draw[red,thick] (2./6,0)--(2./6,2);
    \draw[red,thick] (3./6,0)--(3./6,2);
    \draw[red,thick] (4./6,0)--(4./6,2);
    \draw[red,thick] (5./6,0)--(5./6,2);
    \draw[red,thick] (6./6,0)--(6./6,2);
    \draw[red,thick] (7./6,0)--(7./6,2);
    \draw[red,thick] (8./6,0)--(8./6,2);
    \draw[red,thick] (9./6,0)--(9./6,2);
    \draw[red,thick] (10./6,0)--(10./6,2);
    \draw[red,thick] (11./6,0)--(11./6,2);

    \draw[red,thick] (0,3+1./6)--(2,3+1./6);
    \draw[red,thick] (0,3+2./6)--(2,3+2./6);
    \draw[red,thick] (0,3+3./6)--(2,3+3./6);
    \draw[red,thick] (0,3+4./6)--(2,3+4./6);
    \draw[red,thick] (0,3+5./6)--(2,3+5./6);
    \draw[red,thick] (0,3+6./6)--(2,3+6./6);
    \draw[red,thick] (0,3+7./6)--(2,3+7./6);
    \draw[red,thick] (0,3+8./6)--(2,3+8./6);
    \draw[red,thick] (0,3+9./6)--(2,3+9./6);
    \draw[red,thick] (0,3+10./6)--(2,3+10./6);
    \draw[red,thick] (0,3+11./6)--(2,3+11./6);

    \draw[red,thick] (3+2./6,3)--(3+2./6,5);
    \draw[red,thick] (3+4./6,3)--(3+4./6,5);
    \draw[red,thick] (3+6./6,3)--(3+6./6,5);
    \draw[red,thick] (3+8./6,3)--(3+8./6,5);
    \draw[red,thick] (3+10./6,3)--(3+10./6,5);
    \draw[red,thick] (3,4)--(5,4);

    \draw[red,thick] (3,2./6)--(5,2./6);
    \draw[red,thick] (3,4./6)--(5,4./6);
    \draw[red,thick] (3,6./6)--(5,6./6);
    \draw[red,thick] (3,8./6)--(5,8./6);
    \draw[red,thick] (3,10./6)--(5,10./6);
    \draw[red,thick] (4,0)--(4,2);

    \draw[red,thick] (6,2./4)--(8,2./4);
    \draw[red,thick] (6,4./4)--(8,4./4);
    \draw[red,thick] (6,6./4)--(8,6./4);
    \draw[red,thick] (6+2./3,0)--(6+2./3,2);
    \draw[red,thick] (6+4./3,0)--(6+4./3,2);

    \draw[red,thick] (6+2./4,3)--(6+2./4,5);
    \draw[red,thick] (6+4./4,3)--(6+4./4,5);
    \draw[red,thick] (6+6./4,3)--(6+6./4,5);
    \draw[red,thick] (6,3+2./3)--(8,3+2./3);
    \draw[red,thick] (6,3+4./3)--(8,3+4./3);

  \end{tikzpicture}
  \caption{Evolution of the ratio on a $48\times 48$ grid and twelve
    CPU.}
  \label{fig:ratio}
\end{figure}
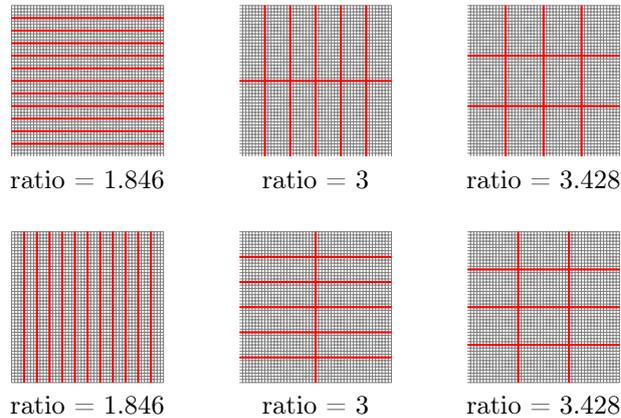

Each CPU communicates the computed flow on the common
boundary to his neighbor. So the more CPU we have, the more important
the communication time is. More, the estimation of the optical
flow is very sensitive at the boundary. For these reasons we want to
optimize the number of pixels at the boundary of each part of the
image. This is why we look for the largest ratio
$\frac{\text{area}}{\text{perimeter}}$ where \emph{area} is the total
number of pixels in the sub-domain and \emph{perimeter} is the number of
pixels on the boundary. In the figure \ref{fig:ratio}, we propose an example of
a $48\times 48$ grid and a decomposition for twelve 
CPU. We give the different values of this ratio with respect to every possible
splittings.
Because the grid is squared in this example, the
ratio is the same for the symmetric decompositions. Finally, we can see
that the ratio is better if we split the image in $4\times 3$ (or $3\times 4$).

Since the optical flow estimation is sensitive at the
boundaries, we use an additive Schwarz method to improve the estimation on the interfaces. This
method requires an overlapping between the subdomains.


\subsection{Model decomposition}

\begin{figure}[H]
  \centering
  \begin{tikzpicture}[scale=1.2]
    \node[anchor=south west,inner sep=0, opacity=0.55] at (0,0)
    {\includegraphics[width=4.8cm,height=4.8cm]{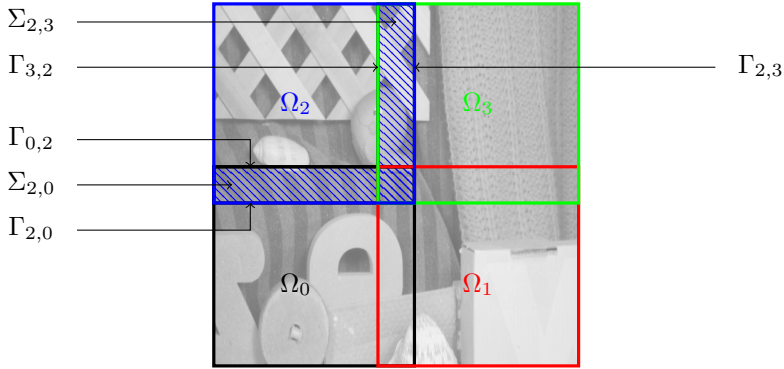}};

    \draw[very thick,pattern=north west lines,pattern color=blue] (0,1.8)
    rectangle (2.2,2.2);
    \draw[very thick,pattern=north west lines,pattern color=blue] (1.8,1.8)
    rectangle (2.2,4); 

    \draw[very thick] (0,0) rectangle (2.2,2.2);
    \draw[very thick,red] (1.8,0) rectangle (4,2.2);
    \draw[very thick,green] (1.8,1.8) rectangle (4,4);
    \draw[very thick,blue] (0,1.8) rectangle (2.2,4);

    \node at (0.9,0.9) {$\Omega_0$};
    \node[very thick,red] at (2.9,0.9) {$\Omega_1$};
    \node[very thick,blue] at (0.9,2.9) {$\Omega_2$};
    \node[very thick,green] at (2.9,2.9) {$\Omega_3$};

    \node at (-2,2) {$\Sigma_{2,0}$};
    \node at (-2,3.8) {$\Sigma_{2,3}$};
    \node at (-2,2.5) {$\Gamma_{0,2}$};
    \node at (-2,1.5) {$\Gamma_{2,0}$};
    \node[very thick] at (6,3.3) {$\Gamma_{2,3}$};
    \node[very thick] at (-2,3.3) {$\Gamma_{3,2}$};

    \draw[->] (-1.5,2)--(0.2,2); 
    \draw[->] (-1.5,3.8)--(2,3.8);
    \draw[->] (-1.5,2.5)--(0.4,2.5)--(0.4,2.2);
    \draw[->] (-1.5,1.5)--(0.4,1.5)--(0.4,1.8);
    \draw[->] (-1.5,3.3)--(1.8,3.3);
    \draw[->] (5.5,3.3)--(2.2,3.3);
  \end{tikzpicture}
  \caption{Example of notations for CPU $i=2$.}
  \label{fig:notation}
\end{figure}

We note $\Omega_i$ the part of the image corresponding to the
$\text{CPU}_i$ and $\mathcal J_i$ the set of all indexes $j$ which are
neighbors to $i$. We define $\Sigma_{i,j}=\Omega_i\cap\Omega_j$ and
$\Gamma_{i,j}=\partial\Sigma_{i,j}\backslash\partial\Omega_j$
(See figure \ref{fig:notation}).

By using
the additive Schwarz method to find an estimation $\tbf{U}_i$ of the
optical flow in the part $\Omega_i$, the problem  
related to (\ref{eq:eqpro1illu}) is 

\begin{numcases}{}
  -\text{div}( \Lambda^k\nabla \tbf{U}^k_i ) + A_{\rho}\tbf{U}^k_i =
  \tbf{F} \text{ in } \Omega_i\nonumber\\
  \diff{\tbf{U}^k_i}{n} = 0 \text{ on
  } \partial\Omega_i\backslash\Gamma_{i,j},\ \forall j\in
  \mathcal J_i \label{schwarz}\\
  \tbf{U}^k_i = \tbf{U}^{k-1}_j \text{ on } \Gamma_{i,j},\ \forall j\in
  \mathcal J_i.\nonumber 
\end{numcases}
Thanks to this method the sequences $(\tbf{U}_i^k)$ converge to
$\tbf{U}_{|_{\Omega_i}}$. The rate of convergence increases with
respect to the size of $\Sigma_{i,j}$. Another convergent method presented by
P.-L. Lions \cite{lions-schwarz-1990} exists with the Robin boundaries conditions.


\subsection{Multi-level parallel method}

To solve the system with the finite element method we use the sofware
\emph{FreeFem++} \cite{freefem}. The default solver of this software for linear systems is 
\emph{UMFPACK} (Unsymmetric MultiFrontal method) which allows to use
non-symmetric matrices. The drawback of this library is that it can only solve
problems smaller than 4GB. It means that we can't treat parts larger than
about $500\times 500$ pixels per CPU so we need to cut the whole image enough in the case of
 large images. This implies to have a large number of CPU which is not
 always the case depending on the machine we work with.

\begin{figure}[H]
  \centering
  \begin{tikzpicture}
    \node[anchor=south west,inner sep=0,opacity=0.65] at (0,0)
    {\includegraphics[width=5.2cm,height=4.8cm]{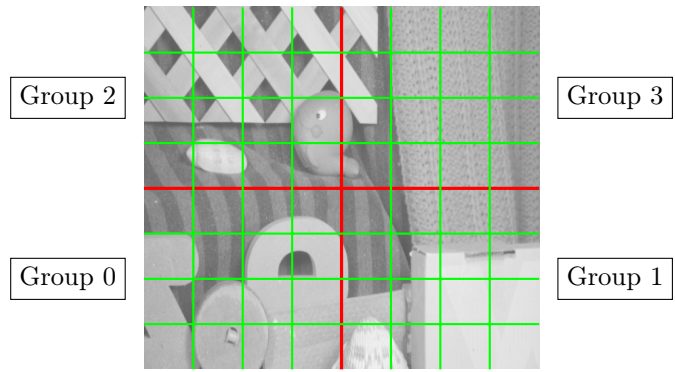}};
    \draw[red,very thick] (0,2.4)--(5.2,2.4);
    \draw[red,very thick] (2.6,0)--(2.6,4.8);
    \foreach \i in {1,2,3,5,6,7}{
      \draw[green, thick] (0,{\i*4.8/8})--(5.2,{\i*4.8/8});      
      \draw[green,thick] ({\i*5.2/8},0)--({\i*5.2/8},4.8);      
    }
    \node[draw] at (-1,1.2) {Group 0};
    \node[draw] at (6.2,1.2) {Group 1};
    \node[draw] at (6.2,3.6) {Group 3};
    \node[draw] at (-1,3.6) {Group 2};

  \end{tikzpicture}
  \caption{Decomposition of an image for four groups of CPU.}
  \label{fig:mumps-ddm}
\end{figure}

To overcome this issue, we use the
\emph{MUMPS} (MUltifrontal Massively Parallel sparse direct
Solver) library which allows the resolution of sparse linear problems in parallel.
The advantage of this solver is that it is not limited by the size of the
problem, so we are able to use high definition
sequences. It also ensures, to some extent, the scalability (see figure
\ref{fig:time-total}).
However, it enforces to apply an LU decomposition to the matrix which can be long if 
this one is large (see figure \ref{fig:diag-bar}).
Combining the domain decomposition method with the use of the \emph{MUMPS}
library (see figure \ref{fig:mumps-ddm}), we can reduce the computation time of
the high definition sequence estimation.

The image is split in the same way as above according to the maximal ratio
between the perimeter and the area of the part. The number of the group in which
belongs the $\text{CPU}_i$ is given by

\begin{equation*}
  \text{group}(\text{CPU}_i) = i\%(\text{nbPart})
\end{equation*}
where the binary operator $a\%b$ states for the remainder of the division of $a$
by $b$ and nbPart is the number of parts in the split image. The rest of the implementation
is the same that the domain decomposition without \emph{MUMPS} except that we act on
a group instead of a single CPU.


\section{Numerical Results}

\begin{figure}[H]
  \centering
  \includegraphics[width=5.2cm]{frame10.png}
  \includegraphics[width=5.2cm]{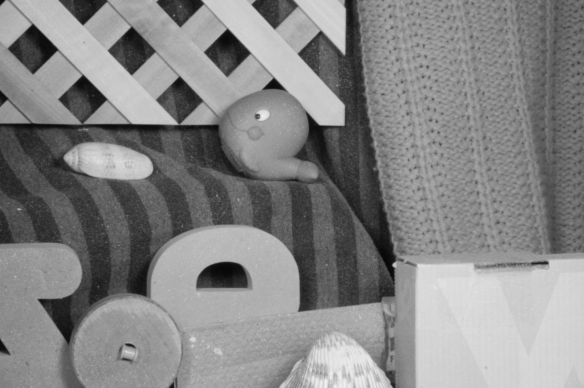}
  \caption{Frames 10 and 11 of the RubberWhale sequence.}
  \label{fig:picture1011}
\end{figure}

To test our algorithms, we use the \emph{RubberWhale} sequence (figure
\ref{fig:picture1011}) given by the Middleburry website at
\url{www.vision.middleburry.edu/flow/}.

\begin{figure}[h!]
  \centering
  \includegraphics[width=5.8cm,height=4.83cm]{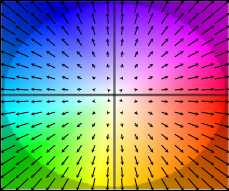}
  \caption{Vector field and its corresponding color map.}
  \label{fig:color-map}
\end{figure}

Following their convention, we
represent the estimated vector field thanks to a color map (figure
\ref{fig:color-map}) which assigns a color to each vector with respect to its
orientation and its norm.

\begin{figure}[H]
  \centering
  \includegraphics[width=5.2cm]{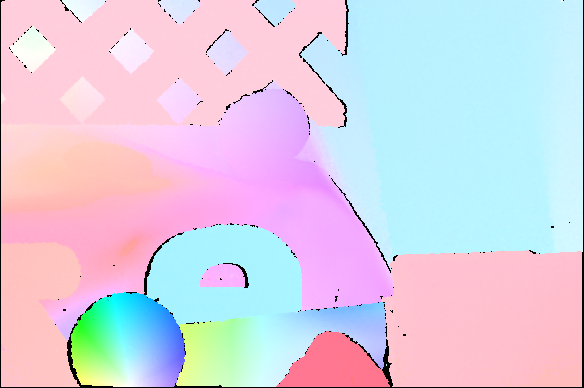}
  \caption{Ground truth solution of the RubberWhale sequence.}
  \label{fig:ground_truth}
\end{figure}

To validate our implementation, we compare our results with the exact solution
also provided by Middelburry (figure
\ref{fig:ground_truth}).

\begin{figure}[H]
  \centering
  \includegraphics[width=7.2cm]{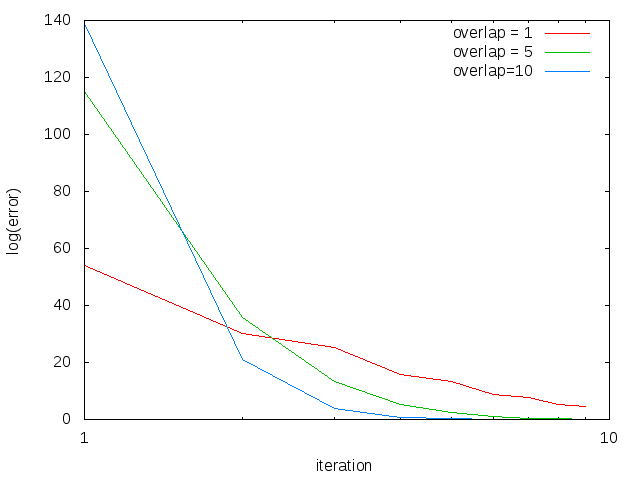}
  \caption{Convergence of the Schwarz method for three different overlaps.}
  \label{fig:schwarz-convergence}
\end{figure}

For optical flow problems, the accuracy of the method is usually evaluated by
computing the \emph{Average Angular Error} (AAE) given by

\begin{equation*}
  AAE = acos\frac{u_{1,h}u_{1,e} + u_{2,h}u_{2,e} + 1}
  {\sqrt{(u_{1,h}^2 + u_{2,h}^2 + 1)(u_{1,e}^2 + u_{2,e}^2 + 1)}}
\end{equation*}
and the \emph{Endpoint Error} (EE) given by

\begin{equation*}
  EE = \sqrt{(u_{1,h}-u_{1,e})^2 + (u_{2,h}-u_{2,e})^2}
\end{equation*}
where $\tbf u_h=(u_{1,h},u_{2,h})$ is an approximation of the vector field and
$\tbf u_e=(u_{1,e},u_{2,e})$ represents the exact optical flow. Using the
resolution of the system (\ref{eq:BW2}), our algorithm reaches an average angular error
equals to 20.89 and an endpoint error equals to 0.38.
In the litterature, the best angular errors go from 1 to 15 and the endpoint
errors from 0.07 to 0.39 for the equivalent evaluation test case
called \emph{Army} (see the evaluation table at
\url{http://vision.middlebury.edu/flow/eval/results/results-a1.php}). There are
two reasons 
to explain that our error is large compared to these values. 
First, we recall that \emph{FreeFem++} \cite{freefem} uses
unstructured meshes and the computation of the angular error is not invariant
with respect to the choice of the mesh. The other reason is that, even if we obtain a good 
approximation of the vector field direction (see the results presented bellow), due to the large 
value of the regularization in the non-egde areas, we under-estimate the vector norms.
There exists some iterative strategy to improve this value but since we are
principally interested in  
improving the computational time and in the adaptation of the regularization
parameter, we don't use it. 
To validate the domain decomposition, we need to verify the convergence of the Schwarz
method which means that

\begin{equation*}
  \mid \tbf U^k - \tbf U^{k+1}\mid \underset{k\to\infty}{\longrightarrow} 0.
\end{equation*}
We have split the image in four parts and tested the impact of the size of the overlap by using three
different values, see figure \ref{fig:schwarz-convergence}.

We can see that the Schwarz method has a fast convergence and that the
speed of convergence increases with the size of the overlap.
However, if the overlap is large, the time of the construction of
the matrix $A$, which is the longest 
part of the code, is large too. Hence, we
need to choose the size of the overlap considering an optimal ratio between the 
rate of convergence and the speed of computation time.
Therefore, we choose an overlap of five pixels in the following
tests.

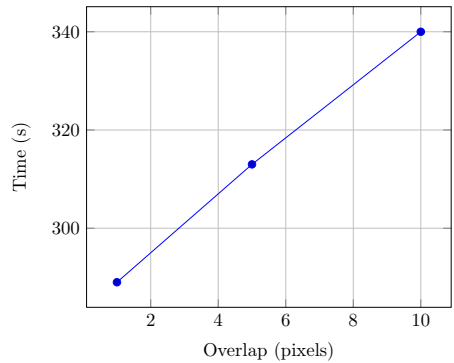
\begin{figure}[H]
  \centering
  \begin{tikzpicture}[scale=0.7]
    \begin{axis}[grid=major, xlabel={Overlap (pixels)}, ylabel={Time (s)}]
      \addplot coordinates {(1,289) (5,313) (10,340)};
    \end{axis}
  \end{tikzpicture}
  \caption{Computation time with respect to the overlap.}
  \label{fig:time-gap}
\end{figure}

\begin{figure}[H]
  \centering
  \subfigure{
    \includegraphics[width=5.2cm]{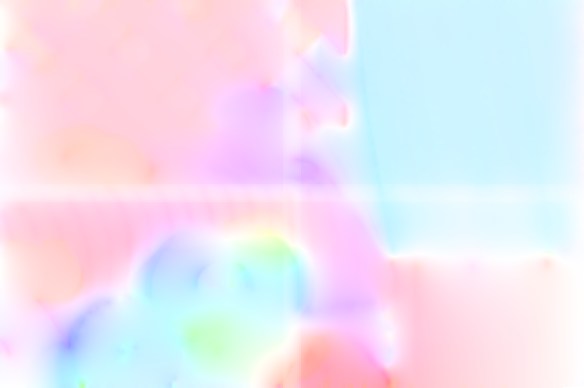}
  }
  \subfigure{
    \includegraphics[width=5.2cm]{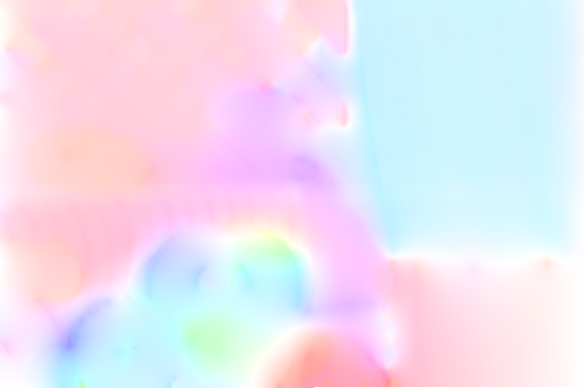}
  }
  \subfigure{
    \includegraphics[width=5.2cm]{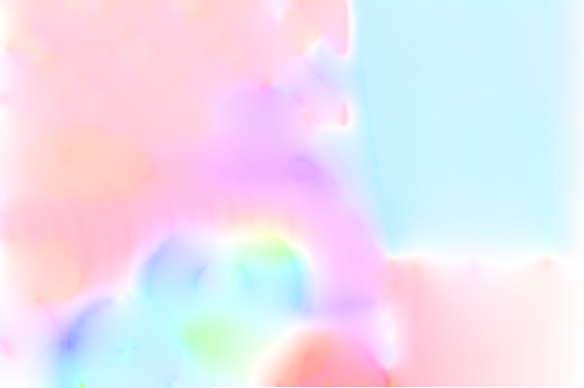}
  }
  \subfigure{
    \includegraphics[width=5.2cm]{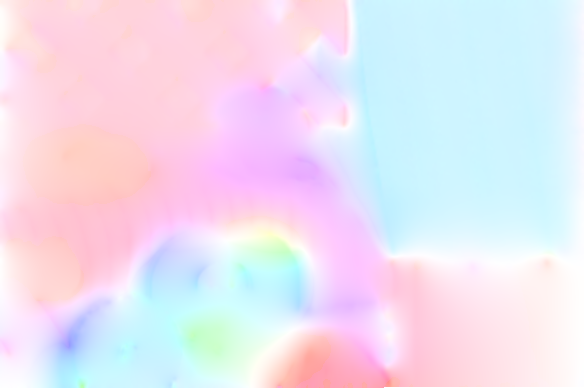}
  }
  \caption{Results for a 2x2 separation, one image per iteration of Schwarz and with
    an overlap of five pixels.}
  \label{fig:evolution}
\end{figure}

There exist other methods without overlapping which may be theoretically
faster \cite{lions-schwarz-1990}. In a forthcoming work, we will consider this class of methods.
However, as we have seen, an 
overlap of 5 pixels is enough, so for large images the additional time can be
neglected.
On the figure \ref{fig:evolution}, we present the evolution of the estimation
according to the iterations of the Schwarz method. We can see that 
since the third iteration, we already have a good estimation of the
solution at the interfaces.

\begin{figure}[H]
  \centering
  \begin{tikzpicture}[scale=0.7]
    \begin{axis}[grid=major, xlabel={Number of parts}, ylabel={Time (s)},
      legend entries={1 CPU Per Part, 4 CPU Per Part}]
      \addplot coordinates {(2,609) (4,270) (8,134)};
      \addplot coordinates {(2,427) (4,224) (8,139)};
    \end{axis}
  \end{tikzpicture}
  \caption{Computation times of different configurations of the multi-level
    parallelism.}
  \label{fig:time-total}
\end{figure}
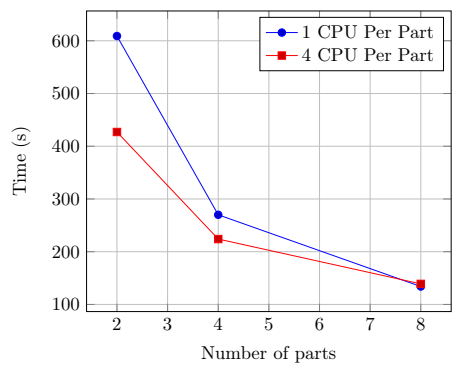

In order to test the implementation of the multi-level parallelism, we have launched
the same test case for
different splittings of the image (2, 4 and 8 parts) and different numbers of CPU
Per Part (CPP). In all cases, we have done ten
iterations of the Schwarz method and we have kept an overlap of five pixels. On
the figure \ref{fig:time-total}, we present the improvement of the computation
times.

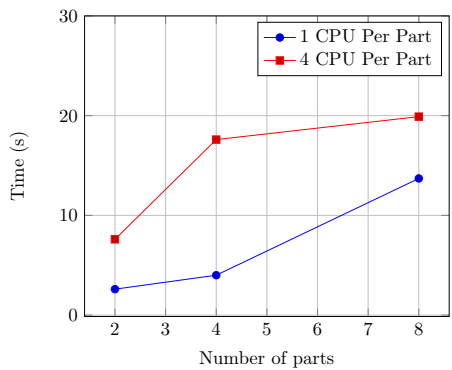
\begin{figure}[H]
  \centering
  \begin{tikzpicture}[scale=0.7]
    \begin{axis}[ymax=30, grid=major, xlabel={Number of parts}, ylabel={Time (s)},
      legend entries={1 CPU Per Part, 4 CPU Per Part}]
      \addplot coordinates {(2,2.6) (4,4) (8,13.7)};
      \addplot coordinates {(2,7.6) (4,17.6) (8,19.9)};
    \end{axis}
  \end{tikzpicture}
  \caption{Communication times with respect to the number of CPU and the splitting (does
    not include the intern communications due to the \emph{MUMPS} solver).}
  \label{fig:time-comm}
\end{figure}

To understand the low efficiency of the multi-level implementation in the case where we use
32 CPU and split with 4 parts, we first represent on the figure \ref{fig:time-comm} the
communication times obtained in the different cases.

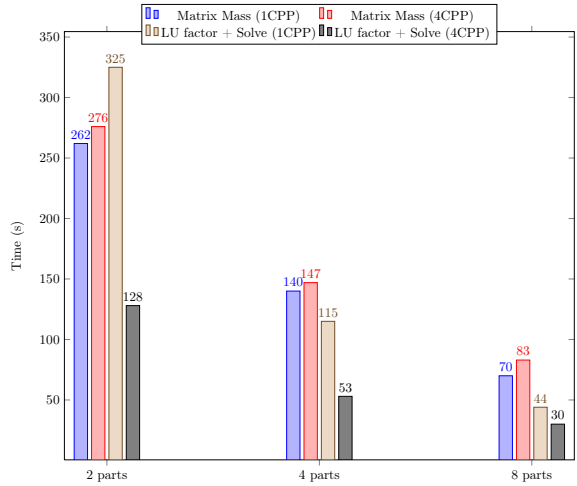
\begin{figure}[H]
  \centering
  \begin{tikzpicture}[scale=0.5]
    \begin{axis}[width=15cm, ybar=3pt, xtick={1,2,3}, xticklabels={2 parts, 4
        parts, 8 parts}, ylabel={Time (s)},
      legend entries={Matrix Mass (1CPP), Matrix Mass (4CPP), LU factor + Solve
        (1CPP),LU factor + Solve (4CPP)},
      legend style={at={(0.5,0.97)},anchor=south},
      legend columns=2,
      nodes near coords]
      \addplot coordinates {(1,262) (2,140) (3,70)};
      \addplot coordinates {(1,276) (2,147) (3,83)};
      \addplot coordinates {(1,325) (2,115) (3,44)};
      \addplot coordinates {(1,128) (2,53) (3,30)};
    \end{axis}
  \end{tikzpicture}
  \caption{Detailed computations times.}
  \label{fig:diag-bar}
\end{figure}

The increase of the communication times is not enough to explain the result
obtained. Indeed, in every parallel implementations, the communication times 
increase with respect to the number of CPU used but it is usually balanced with
the time saved in the computations. So, in order to give more details, we
present in the figure \ref{fig:diag-bar} the times of
the two main parts of the computation: the construction of the mass matrix and
the LU factorization with the resolution part. 
We can see
that adding more CPU to a part slightly increases the time of construction of the
mass matrix. However, it allows to decrease the time of the LU factorization. On
our architecture, if we split the images into several parts, the main part of the
computation is the mass matrix construction. If we use a large number of CPU per
part, we reduce the difference between the time saved in the LU decomposition and
the time lost in the matrix construction. More, we increase the communication times
and
the \emph{MUMPS} parallel solver doesn't balance that. 

\begin{figure}[H]
  \centering
  \subfigure{
    \includegraphics[width=5.2cm]{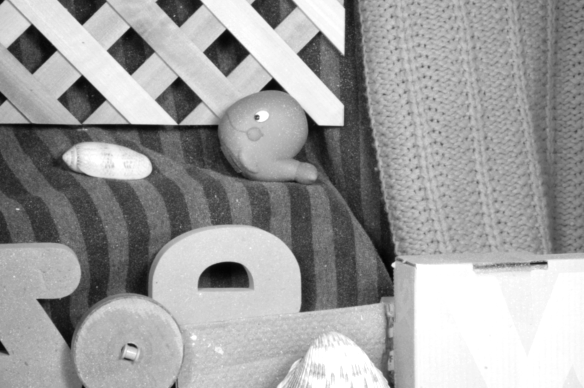}
  }

  \caption{Modified first frame of the RubberWhale sequence for the varying
    illumination test case.}
  \label{fig:mod-seq}
\end{figure}

To evaluate the accuracy of the computed flow in the case of varying
illumination, we use the RubberWhale test case where we have modified the first
frame by increasing the brightness intensity (see figure \ref{fig:mod-seq}).
Since we haven't modified 
the motion between the two pictures, the exact solution is the same that in the
previous test case. 

\begin{figure}[H]
  \centering

  \subfigure{
    \includegraphics[width=5.2cm]{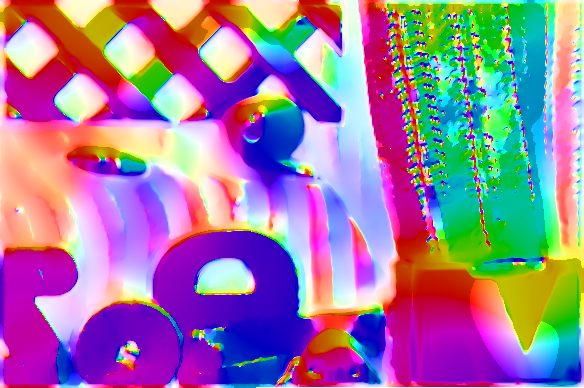}
  }
  \subfigure{
    \includegraphics[width=5.2cm]{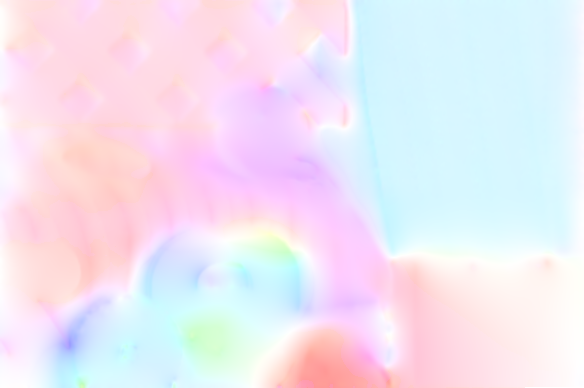}
  }
  \caption{Left: Estimation with (right) and without (left) treatment of the varying illumination. The
    images were split in four parts.}
  \label{fig:illum}
\end{figure}

On the figure \ref{fig:illum}, we present the result obtained with and without
the treatment of the
brightness variation. On the right hand side, we can see that the solution
without treatment is not well estimated. On the other hand, the model which uses the modified
assumption (\ref{eq:OF2}) allowing the brightness variation gives a much accurate
optical flow still well oriented.

\begin{figure}[H]
  \centering
  \includegraphics[width=5.2cm]{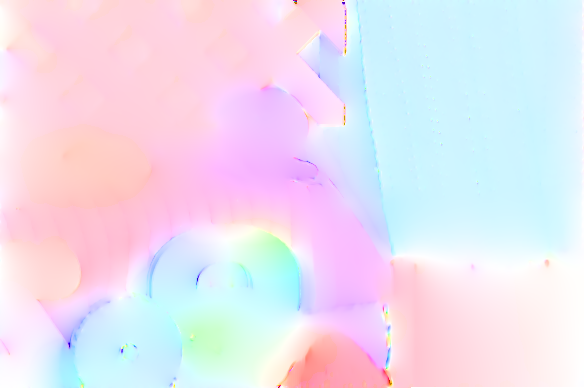}
  \caption{Estimated optical flow after ten
    iterations of adaptation (result for a 2x2
    separation).}
  \label{fig:regu-alpha}
\end{figure}

The next test (figure \ref{fig:regu-alpha}) consists of evaluating the efficiency of the
adaptive regularization. The adaptation steps are done with the
Schwartz iterations. We obtain a better definition of the edges.

\begin{figure}[H]
  \centering
  \includegraphics[width=5.2cm]{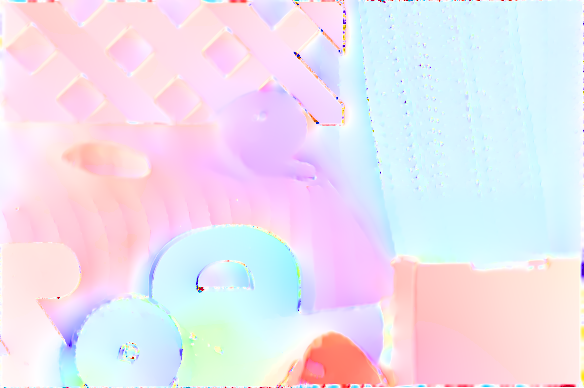}
  \caption{Estimated optical flow after ten
    iterations of adaptation for the test with
    varying
illumination (results for a 2x2 separation).}
  \label{fig:regu-alpha-illum}
\end{figure}

On the figure \ref{fig:regu-alpha-illum}, we can see that the adaptive
regularization is still working for the test with non-constant brightness. In
this case, the edges are even better defined.

\begin{figure}[H]
  \centering
  \subfigure{
    \includegraphics[width=5.2cm]{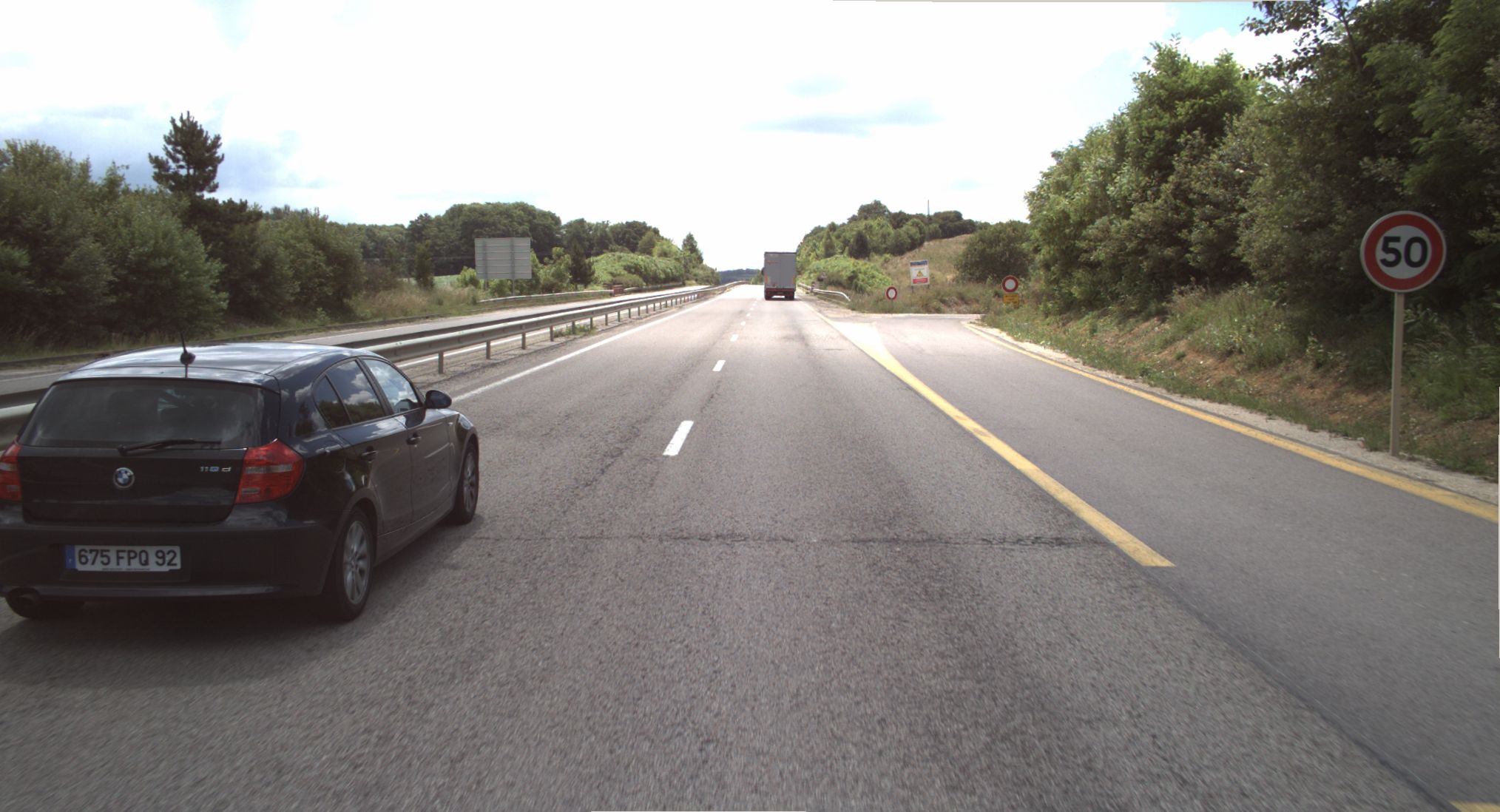}
  }
  \subfigure{
    \includegraphics[width=5.2cm]{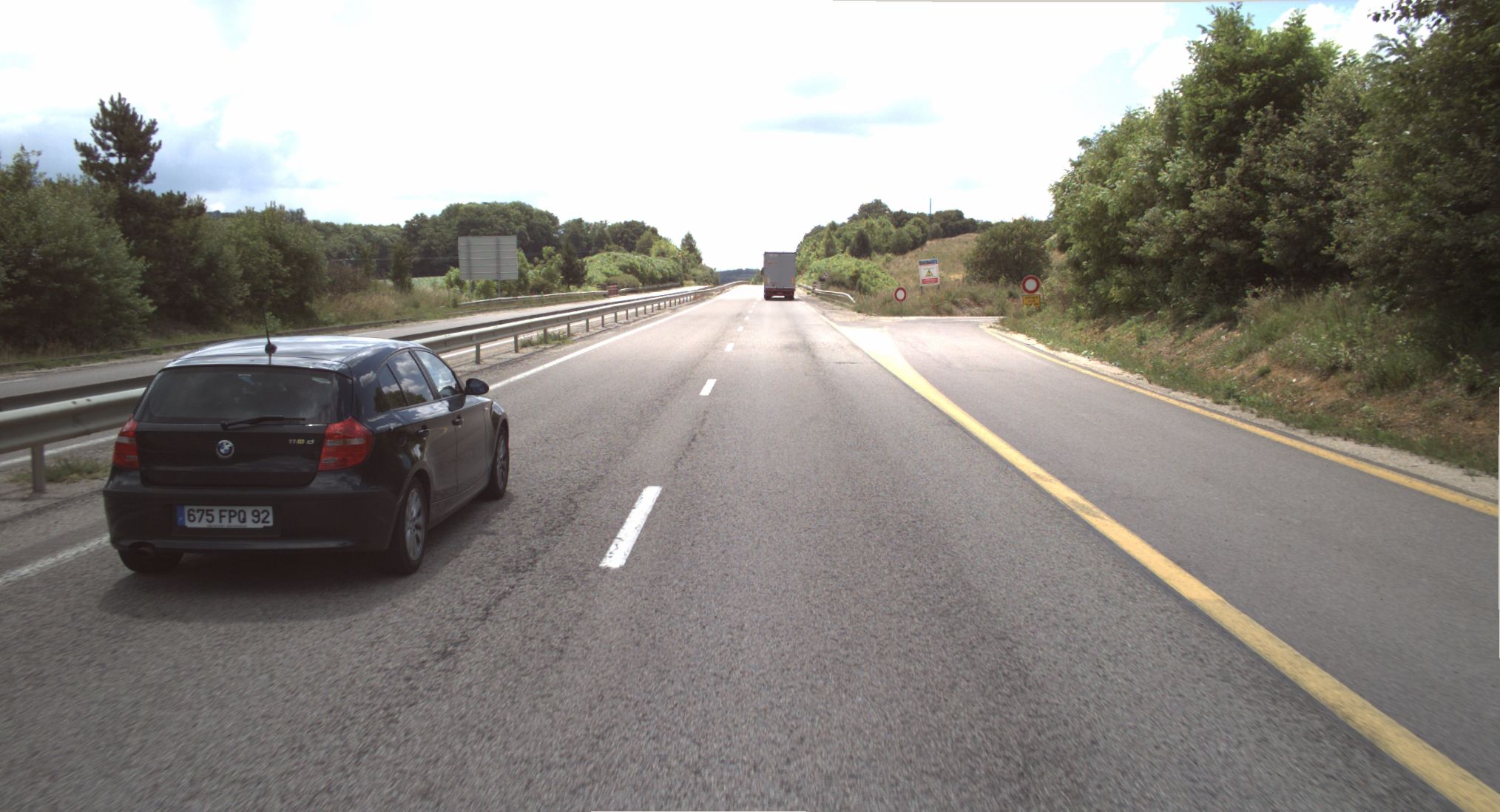}
  }

  \subfigure{
    \includegraphics[width=5.2cm]{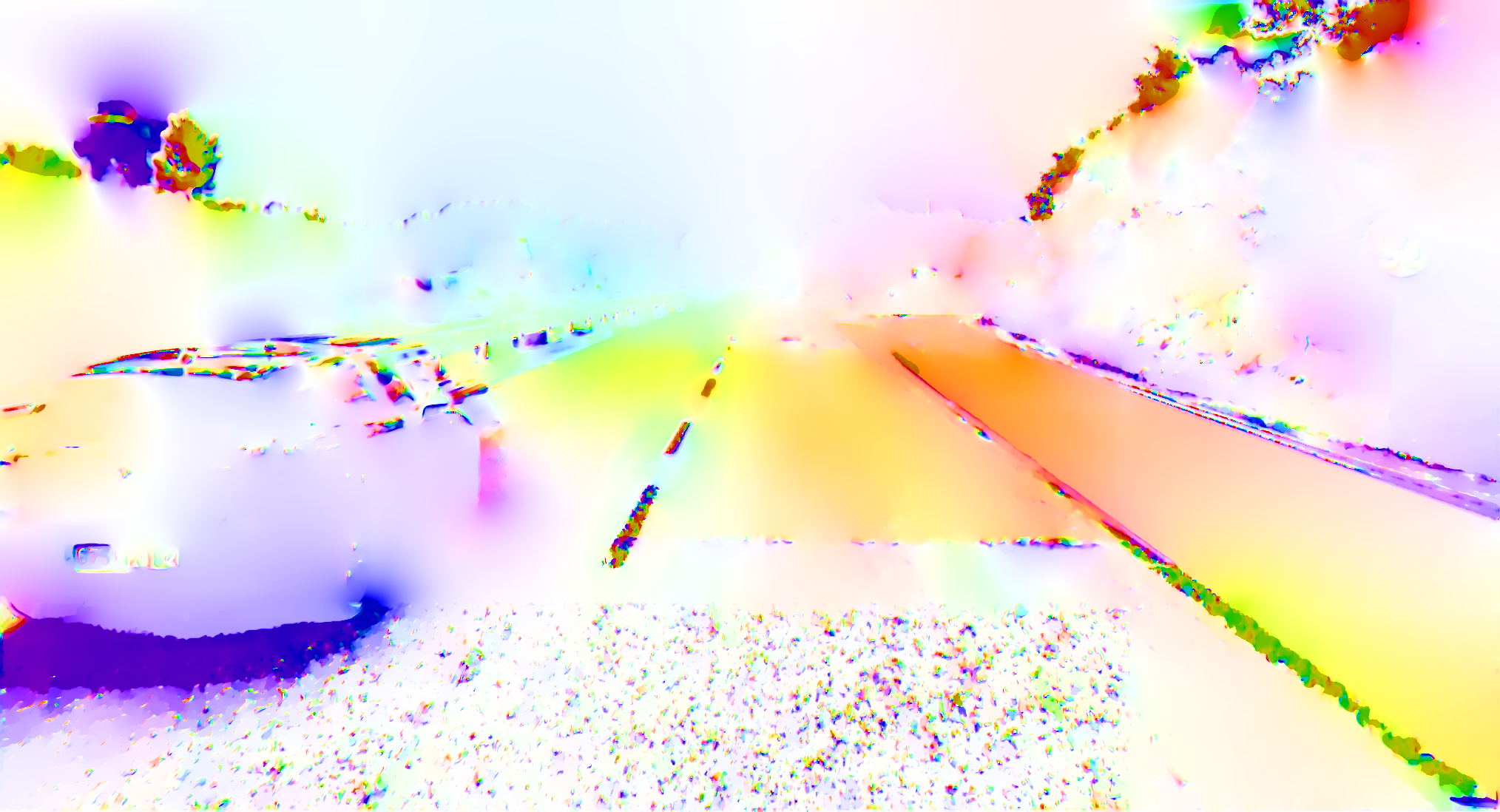}
  }
  \subfigure{
    \includegraphics[width=5.2cm]{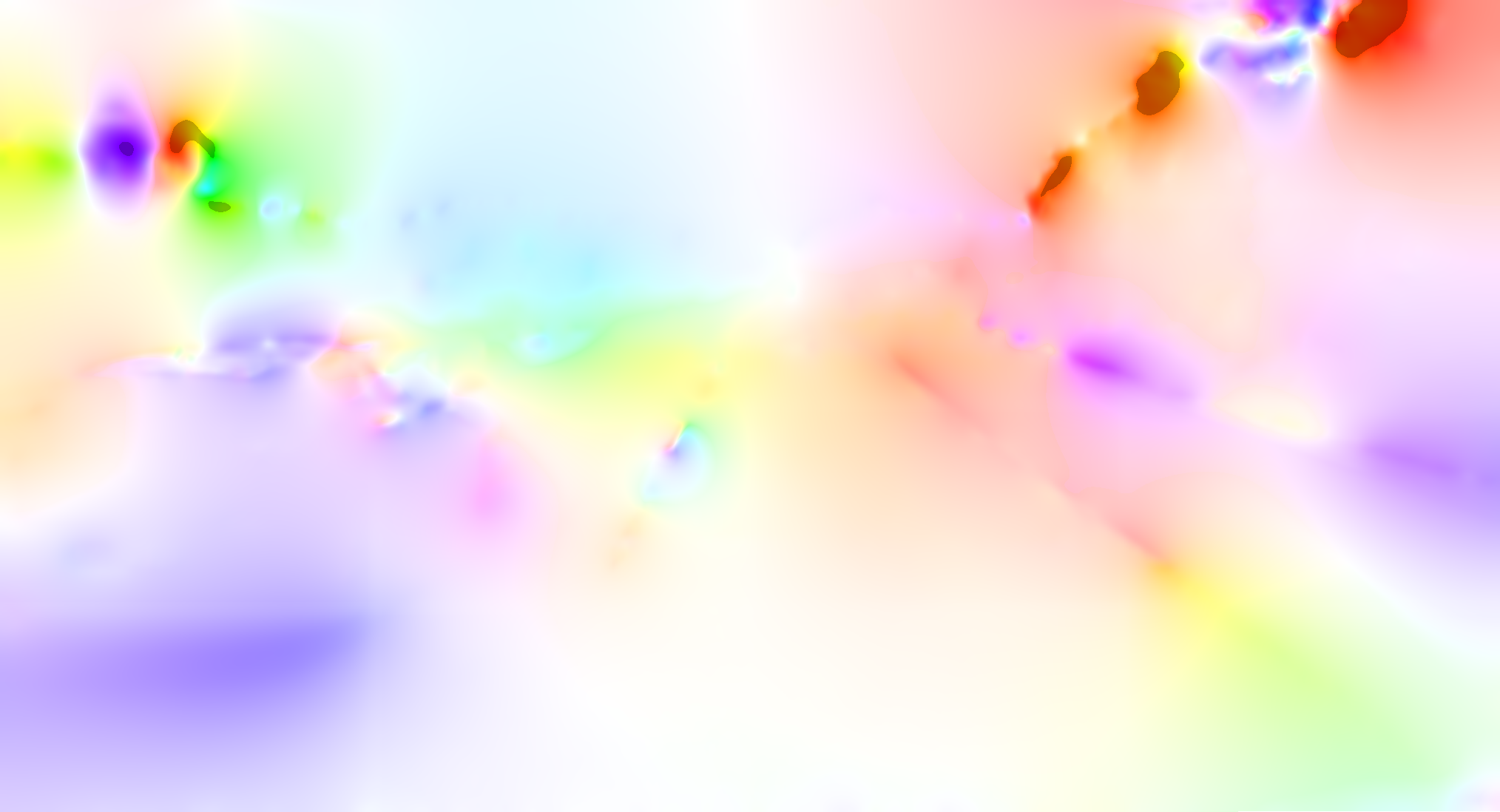}
  }
  \caption{Up: Test case of the Highway sequence given by the Cerema. Bottom: Optical flow
    estimation obtained with (right) and without (left) treatment of the varying
    illumination. The images were divided in sixteen parts.}
  \label{fig:voiture}
\end{figure}

\begin{figure}[!h]
  \centering
  \subfigure{
    \includegraphics[width=5.2cm]{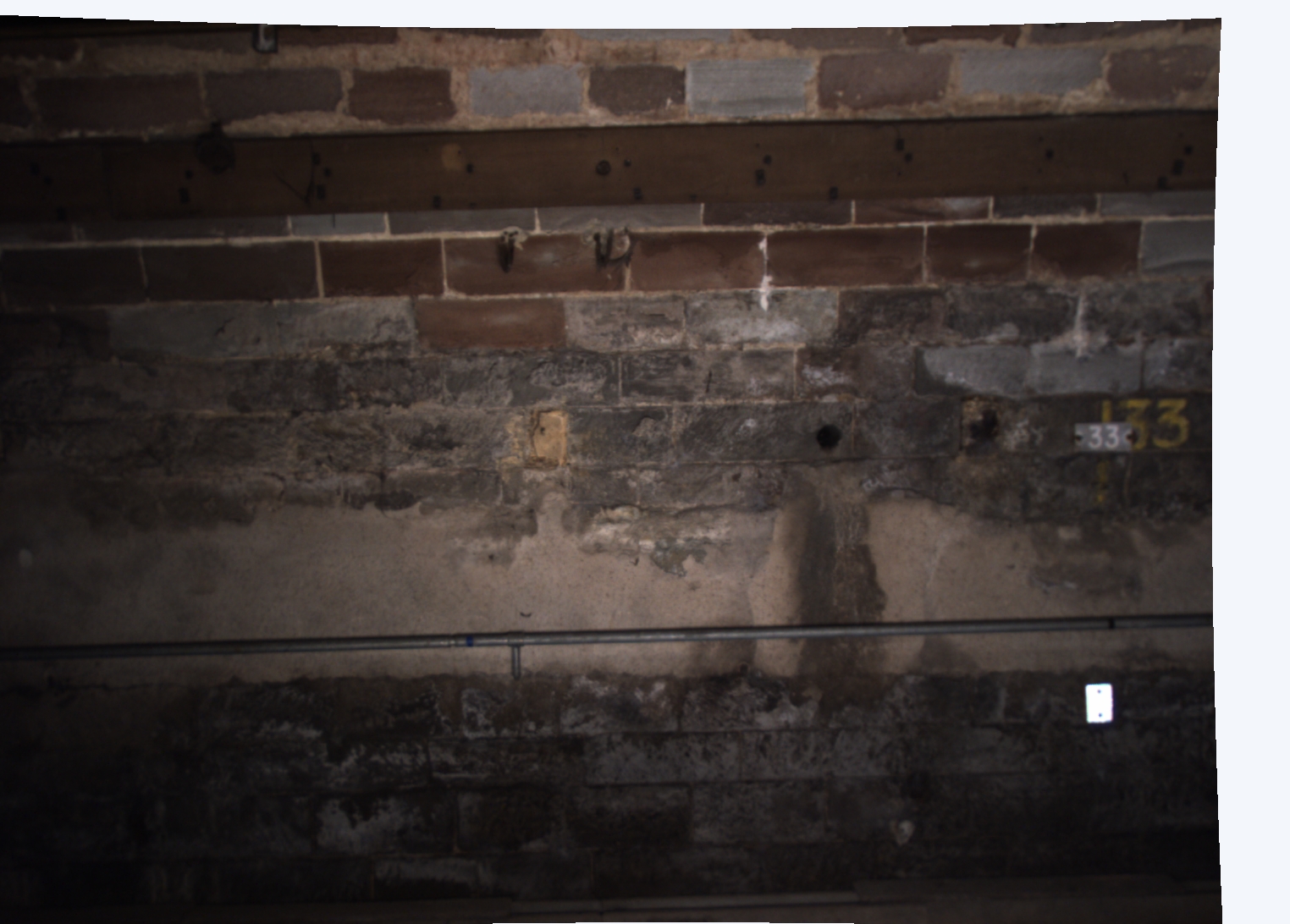}
  }
  \subfigure{
    \includegraphics[width=5.2cm]{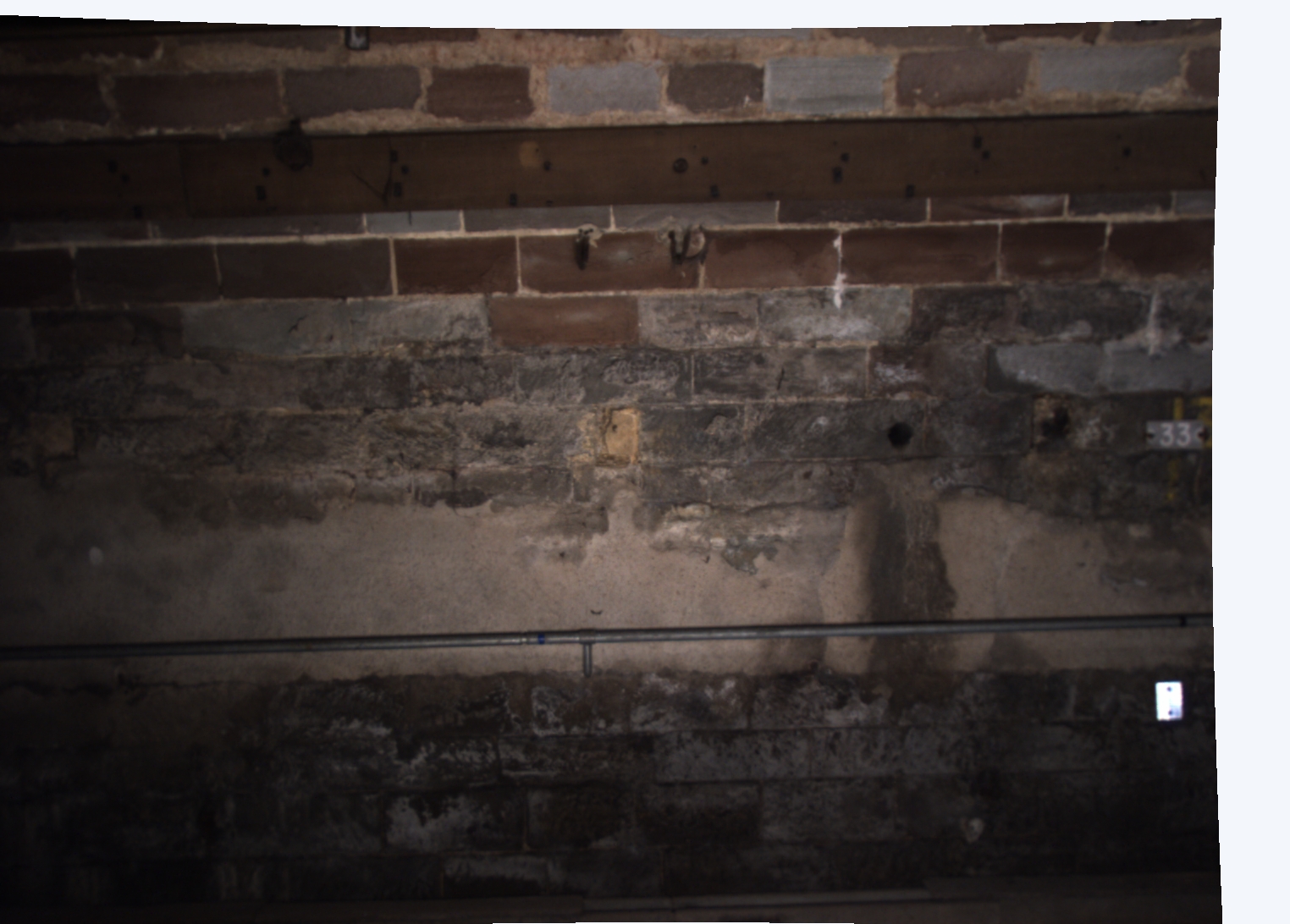}
  }

  \subfigure{
    \includegraphics[width=5.2cm]{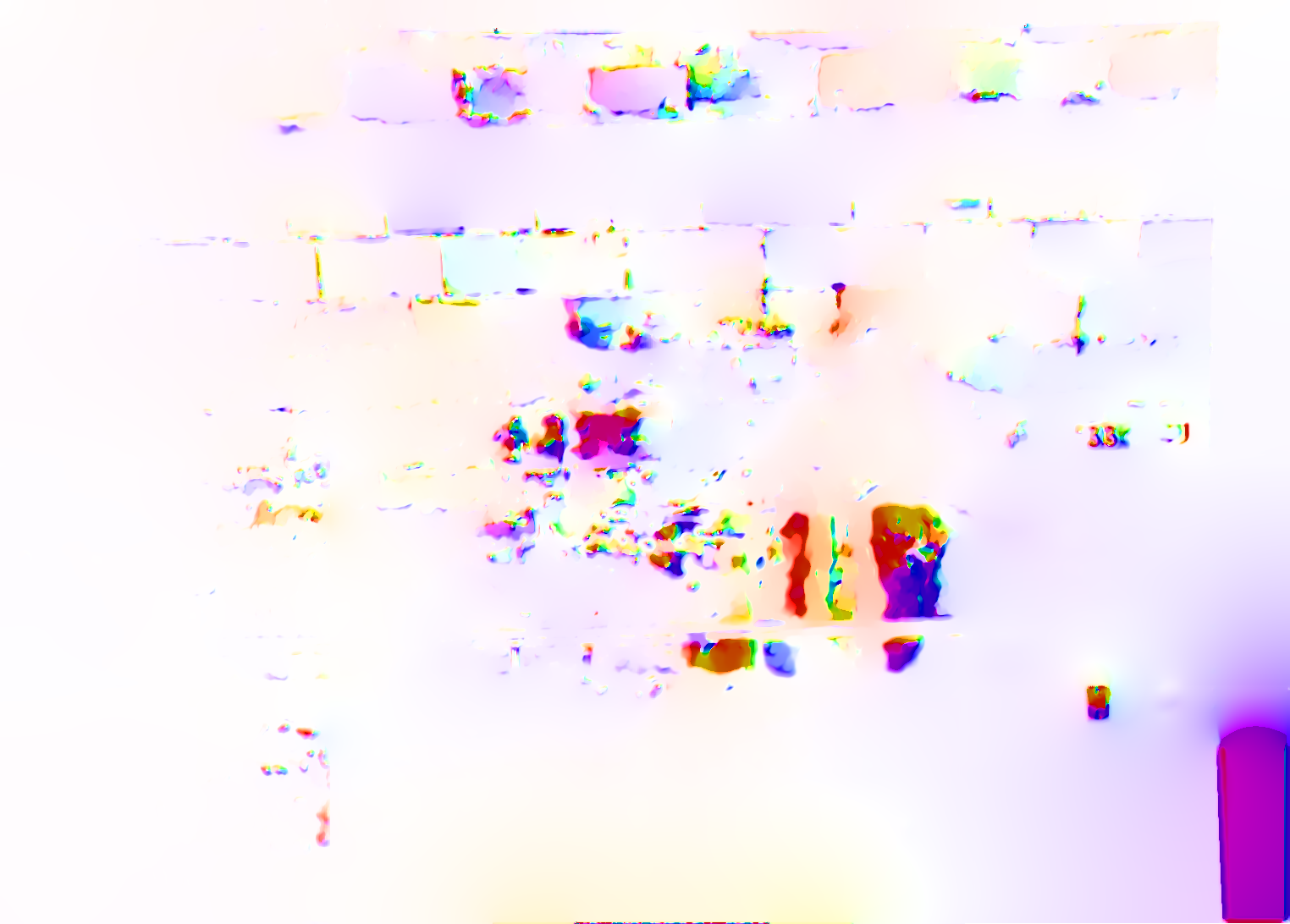}
  }
  \subfigure{
    \includegraphics[width=5.2cm]{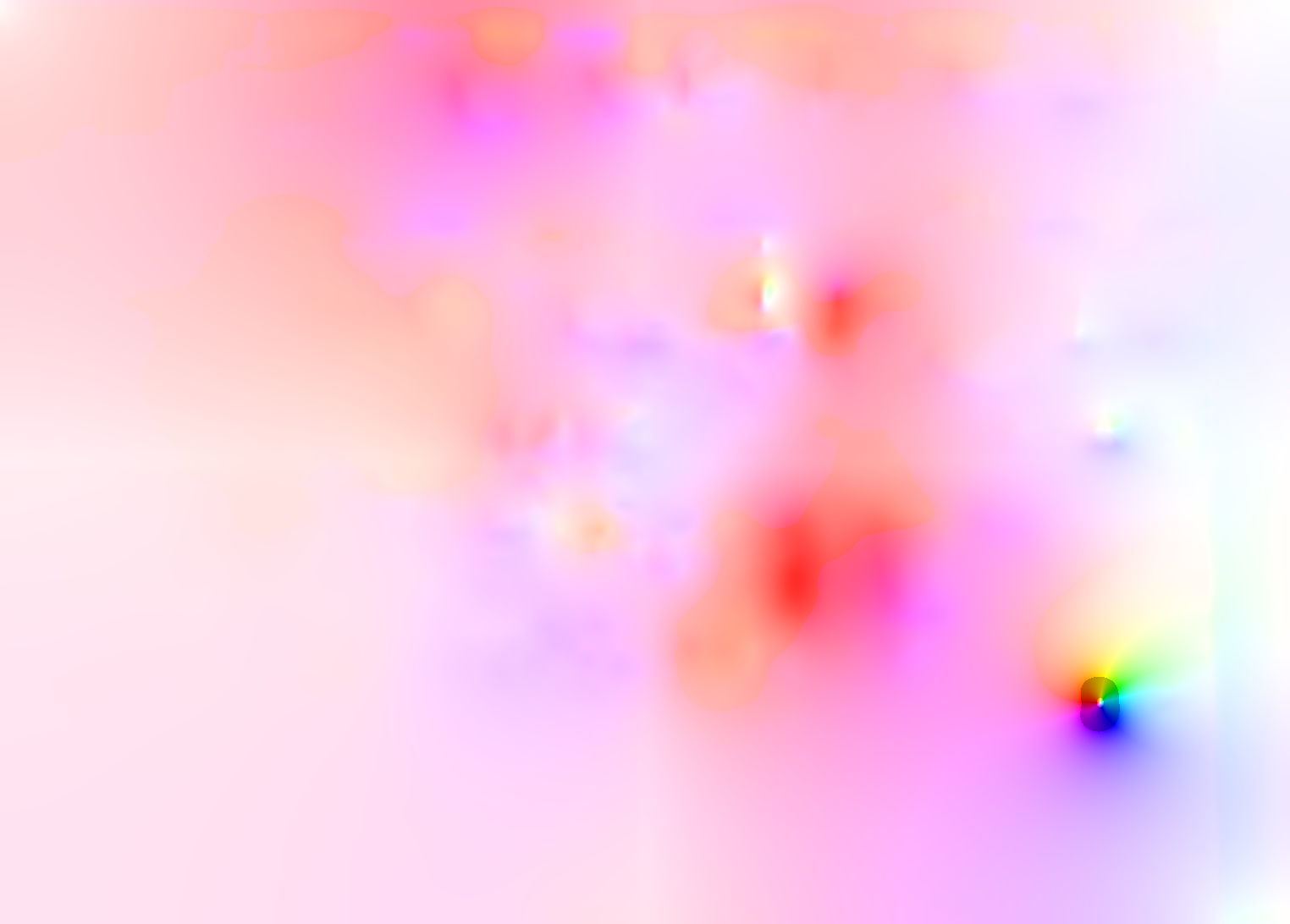}
  }
  \caption{Up: Test case of the Wall sequence given by the Cerema. Bottom: Optical flow
    estimation obtained 
with (right) and without (left) treatment of the varying
    illumination. The images were divided in sixteen parts.}
  \label{fig:mur}
\end{figure}

Finally, on the figures \ref{fig:voiture} and \ref{fig:mur}, we have used two sequences
of real images provided by the \emph{Centre d'études et d'expertise sur les
  risques, l'environnement, la mobilité et l'aménagement} (Cerema). These pictures present 
two main interests. First, the high resolution. We have $2028\times 1098$ pixels for the highway
sequence and $1524\times 1092$ pixels for the wall sequence which corresponds to a
vector field of about two billions of pixels to determine. The second interest
is that the brightness and the texture are natural.\\

The highway sequence presents a non-constant brightness with a more
complex texture and a lot of occlusions as the
white lines on the road or the large motion of the car. We can see on the figure
\ref{fig:voiture}, the difference between the model with and without 
the varying illumination. On the left hand side, we have the solution without
treatment. Again, we see that 
this is not correctly estimated because of the variation of the natural
lighting. On the right, despite the occlusion areas which could be improved
with a large displacement algorithm, we obtain a better approximation. 

The figure \ref{fig:mur} represents a wall of a tunnel. On this sequence the
motion to estimate is linear. The 
challenge is to deal with irregular textures. 
This figure shows that our method is still efficient in this case. Indeed, the solution
is quite smooth and linear except for the white square (bottom
right) which may be improved by using a large displacement algorithm. This
approach will be implemented in a further work.\\

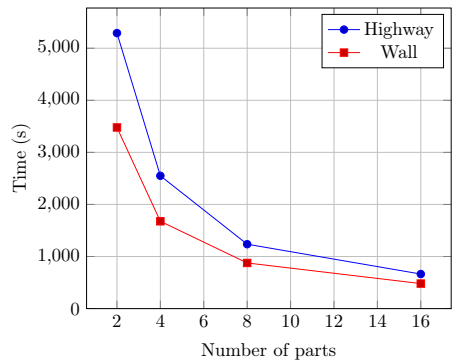
\begin{figure}[H]
  \centering
  \begin{tikzpicture}[scale=0.7]
    \begin{axis}[grid=major, xlabel={Number of parts}, ylabel={Time (s)},
      legend entries={Highway, Wall}]
      \addplot coordinates {(2,5291) (4,2550) (8,1236) (16,663)};
      \addplot coordinates {(2,3478) (4,1676) (8,876) (16,480)};
    \end{axis}
  \end{tikzpicture}
  \caption{Computation times of the high definition tests for different splittings.}
  \label{fig:time-total2}
\end{figure}

In the figure \ref{fig:time-total2}, we give the computation times obtained with
the two high definition tests. We recall that the pictures have about two billions of
pixels and were not reduced. As in the previous cases, we perform ten iterations
of the Schwarz algorithm.

\section*{Conclusion}

In this article we have used the finite element method to solve the optical flow
problem with varying illumination. Thanks to the additive Schwarz method, we have implemented the domain
decomposition in order to parallelize the computations. We have shown that this method
is an efficient way to decrease the computation time and to handle high
resolution sequences.  
We have also used a second level of parallelism in order to reduce the execution
time even more and we have  
shown that such a massively parallel approach yields to an important gain of time. 

The finite element method has also permitted to use an adaptive strategy of the regularization
parameter, hence to efficiently combine the quality of the optical flow estimation
with a method that preserves  
the edges and the fine features of the computed flow.\\


\textbf{Acknowledgments:} The authors would like to thank the \emph{Centre d'etudes et d'expertise sur les
  risques, l'environnement, la mobilité et l'aménagement (Direction territoriale
  Est, laboratoire de Strasbourg)} for their suggestions and their data.

\nocite{bruhn-clg-2005}
\nocite{barron-smooth-1994}
\nocite{memin-multigrid-1998}
\nocite{schnorr2}
\nocite{weickert_complementary}
\nocite{weickert2}

\bibliographystyle{plain}
\bibliography{references_gilliocq}

\end{document}